\documentclass[letterpaper, 10 pt, conference]{ieeeconf}  

\IEEEoverridecommandlockouts                              

 \usepackage{graphics} 
\usepackage{times} 
\usepackage{amsmath} 
\usepackage{amssymb}  
 \usepackage{color}
\usepackage{bm}
\usepackage{array,multirow,graphicx}
\usepackage{amsfonts}
\usepackage[]{algorithm2e}
\usepackage{multicol}
\usepackage{scalerel,stackengine}
\stackMath
\newcommand\reallywidehat[1]{%
\savestack{\tmpbox}{\stretchto{%
  \scaleto{%
    \scalerel*[\widthof{\ensuremath{#1}}]{\kern-.6pt\bigwedge\kern-.6pt}%
    {\rule[-\textheight/2]{1ex}{\textheight}}
  }{\textheight}%
}{0.5ex}}%
\stackon[1pt]{#1}{\tmpbox}%
}
\usepackage[citecolor=green,pagebackref=true]{hyperref}

\title{\LARGE \bf
Vision-based deep execution monitoring
}

\author{Francesco Puja, Simone Grazioso, Antonio Tammaro, Valsmis Ntouskos, Marta Sanzari, Fiora Pirri
\thanks{The authors are with ALCOR Lab, DIAG,
        Sapienza University of Rome, Italy
        {\tt\small \{puja, grazioso, tammaro, ntouskos, sanzari, pirri\}@diag.uniroma1.it}}%
}

\begin{document}

\maketitle
\thispagestyle{empty}
\pagestyle{empty}

\begin{abstract}
Execution monitor of high-level robot actions can be effectively improved by visual monitoring the state of the world in terms of preconditions and postconditions that hold before and after the execution of an action. Furthermore a policy for searching where to look at, either for verifying the relations that specify the pre and postconditions or to refocus in case of a failure, can tremendously improve the robot execution in an uncharted environment. It is now possible to strongly rely on visual perception in order to make the assumption that the environment is observable, by the amazing results of deep learning.  In this work we present visual execution monitoring for a robot executing tasks in an uncharted Lab environment. The execution monitor interacts with the environment via a  visual stream that uses two DCNN for recognizing the objects the robot has to deal with and manipulate, and a non-parametric Bayes estimation to discover the relations out of the DCNN features. To recover from lack of focus and failures due to missed objects we resort to  visual search policies via deep reinforcement learning.
\end{abstract}

\section{Introduction}\label{sec:introduction}
Robot perception has been amazingly  boosted in recent years by deep learning  results on object recognition \cite{krizhevsky2012,ren-2015}, relations recognition \cite{Lu-2016}, visual question-answering (VQA)\cite{antol-2015}, activity recognition \cite{Al-Omari-2016}, image annotation \cite{feng-2016}, and  navigation \cite{mirowski2016,zhu2017} among other perceptual based abilities \cite{lecun2015}. The upshot is that  robots interaction with the real world nowadays seems to be in reach.
The phenomenal success of deep learning  across various disciplines motivates us to investigate whether useful representations in the robot task execution monitoring, in unknown environments, can be learned as well.
 
\begin{figure*}[t!]
 \centering
  \includegraphics[width=16cm,height=8cm,]{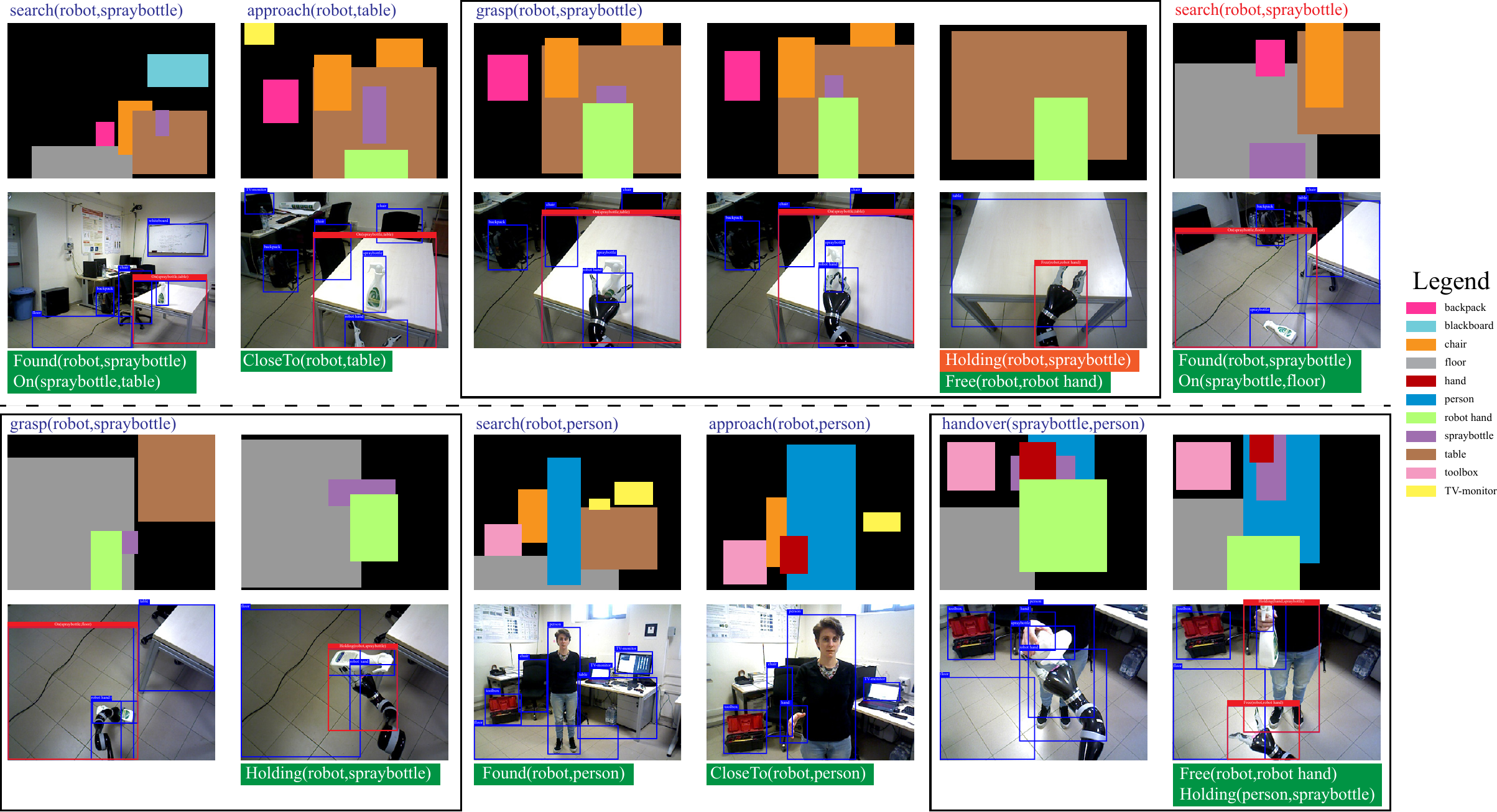}\\
 \caption{Visual execution monitoring (VExM) of the task {\em take the Spraybottle and hand  it  to Person}. Actions written in blue are provided by the planner, while those in red are generated by the visual search policy once a failure is detected. The post-conditions are shown below the images, those not verified, leading to a failure, are shown in red. The second and forth rows of images show the $bb$s of the objects and in red their relations discovered by the visual stream. The first and third rows of images represent the mental maps, here depth within the colored  boxes is not visible. The hand misses the spraybottle (5th image),  postcondition of action grasp  $Holding(robot,spraybottle)$ fails while $Free(robot,robotHand)$ is verified.  After a  visual search the spraybottle is found on the floor. The involved relation in the parsed plan is updated from $On(spraybottle, table)$ to $On(spraybottle,floor)$ and the failed action is re-executed. A further visual search  finds the subject, and finally  the spray bottle is  handed to her. Note the verification of the relations $CloseTo(robotHand,hand)$ and $Free(robot,robotHand)$, this last  ensuring task goal success.}\label{fig:overview}
\end{figure*}

In this paper we address {\em visual execution monitoring} (VExM) in a dynamic uncharted lab environment, for high-level tasks. In order to monitor the state of execution we  embed real time recognition of objects and relations  which we call the {\em visual stream},  within  a hybrid planning model. The visual stream is defined by a hierarchical model combining two deep convolution neural networks (DCNN)s \cite{ren-2015} for object recognition, and a non-parametric Bayes model DPM\cite{ferguson-1973} that uses the active features of the two DCNNs to segment  the depth images collected during task execution. The combination of segmented depth and object labels allows to infer visual relations from the robot point of view. The hybrid planner, on the other hand,   blends deterministic planning, with durable actions \cite{mcdermott-1998,helmert-2009,cashmore2015}, with a contextually optimal {\em visual search} policy inferred from current state execution trained with  DNN \cite{mnih2015,mnih2016}, to cope with both task failures and loss of focus on the task.  For policies inference the VExM  builds states representations of the scene out of the visual stream results, which we call the {\em mental maps}, see Figure \ref{fig:overview}. Indeed,  mental maps are used  for both training the {\em visual search} policies and to recover  focus on the current task,   basing on deep reinforcement learning \cite{mnih2015,mnih2016}.    An overview of a typical task execution in the proposed framework is given in Figure \ref{fig:overview}.

Though visual-based view of robot execution has been somehow faced previously by \cite{lenz2015,Al-Omari-2016,vongbunyong2016,ZhuGKFFGMF17,mirowski2016,zhu2017}, the proposed approach is completely novel in  particular in terms of execution monitoring.

In this work perception is egocentric and we are considering high-level tasks, hence we are not considering motion control, 3D object pose estimation, nor  navigation issues,  we assume that all these problems are tackled by appropriate algorithms, see e.g. \cite{kaiser2016,runz2017,schwarz2015,choi2016,hornung2013,mirowski2016}. 

In summary, we address visual execution monitoring   as an hybrid deterministic/nondeterministic state machine streaming perceptual information, for both monitoring the execution and suitably directing  visual perception. The VExM   refocuses and  redeems from a failure according to learned policies.  Assuming that the visual stream provides a fully observable environment the VExM relies on learned policies \cite{mnih2016} for focusing on the important objects involved in the task execution, and likewise  to assess the discrepancy between the inferred state and the perceived one. Hence in case of failure the VExM can always  resort to  a recovery policy that ensures  an optimized visual search to redeem into a state where execution can be retrieved.

\typeout{-------------------- RELATED -----------------------}
\section{Related Work}\label{sec:relwork}

The earliest definitions of execution monitoring in nondeterministic environments have been introduced in \cite{fikes1971,nilsson1973}. Since then an extraordinary amount of research has been done  to address  the nondeterministic response  of the environment  to robot actions. Several definitions of execution monitoring are reported in \cite{pettersson2005}. For high level robot tasks a review of these efforts is given in \cite{ingrand2017}. The role of perception in execution monitoring was already foreseen in the work of \cite{doyle1986}, likewise recovery from  errors that could occur at execution time was already faced by \cite{wilkins1985}.
Despite this foresight, the difficulties in dealing with scene recognition have directed the effort toward models managing the effects of actions  such as  \cite{sutton1998,bertsekas1995},  allowing to execute actions in partially observable environments \cite{boutilier2000}.   The integration of observations in high level monitor has been recently addressed by several authors, among which  \cite{hornung2014,mendoza2015}. Still, the breakthrough,  is achieved with DCNN and in general with deep learning  for perception \cite{guadarrama2013},   visual planning \cite{ZhuGKFFGMF17} and with deep Reinforcement Learning (RL) \cite{mnih2015,mnih2016}. 

Still for  relations recognition, despite there is a great number of contributions \cite{guadarrama2013,Lu-2016,das2016} for 2D images, it is missed an appropriate method for robot execution. Hence we introduce a form of domain adaptation of the 2D images taken from ImageNet to  segment the 3D images collected by the robot, basing on a DPM \cite{ferguson-1973} that uses the active features of the recognized objects involved in the relation.

Another relevant aspect is visual search and   the environment representation. Since the work of \cite{mnih2015} a wealth of research has been done to make robust and extend RL \cite{bertsekas1995,sutton1998} to very large set of actions, in so developing a new amazing research area, deep RL \cite{mnih2016,levine2016}. The only difficulty is the need to simulate the environment to perform the required huge collection of  experiments. Hence a certain amount of research has been devoted to create new simulation environment such as \cite{mottaghi2016,zhu2017}. Here we contribute with a quite simple representation of the environment using the mental maps generated directly during robot experiments, basing on the depth segmentation and the bounding boxes of the recognized objects.

\typeout{--------------------  OVERVIEW  -----------------------}
\section{Overview of our approach to visual execution monitoring}\label{sec:overview}
The robot execution monitor (VExM) we consider interacts both with the planning environment (described in Section \ref{sec:planenv}) and with the real world environment via the visual stream (Section \ref{sec:visualstream}). It exchanges information about the current state, the executed action  and  the action to be executed in such a state. This information concerns objects, terms and relations mentioned in the pre and postconditions of the action, and it is oriented by the visual search policies (Section \ref{sec:planenv}).

The planner environment has a deterministic component that infers the list of actions, pre and postconditions before execution starts, and a stochastic meta action which is the visual search. Since each of the plans forming a task  has only one action affecting the world (see Section \ref{sec:planenv}) the beforehand inference of the specific plan goal is fine, because this is the only action that can fail, under our hypothesis of not considering robot control, which would lead to a continuous domain. 

{\bf Example} The task is illustrated in Figure \ref{fig:overview}.  From a lexical component not described here, the task is parsed into a number of goals such as  $holding(robot,sb)$, $closeTo(robotHand,hand)$. For each of these goals there is a plan issuing a sequence of actions, but for the visual search, which is  a policy sequencing actions like {\em look-Up, look-Down, turn-Left, turn-Right}. Policies are learned basing on the actor critic paradigm \cite{bertsekas1995,sutton1998},   exploiting the deep learning approach of \cite{mnih2016}.

 At each state of the robot execution of the sequence of actions, the  VExM verifies via the visual stream  the feasibility of the next action visually assessing the relations mentioned in the  action pre and postconditions specified in the state.  
The VExM uses this information to recompute transitions, according to a loss function that averages the discrepancy between the state (pre and postconditions of actions) inferred by the planner and the one observed by the visual stream (see Section \ref{sec:monitor}). 

If the objects of interest are out of the field of view some of the pre or postconditions will fail and the VExM appeals to the visual search to refocus on the object of interest. 
Typical examples are when the robot  misses to position itself in front of a table or when it grasps something which falls down (see Figure \ref{fig:overview}). In these cases the visual search has to output the nice sequence of actions bringing   the missed object back in the visual field.

In summary,  with the VExM we do a step  toward bridging the robot language and behavior between symbolic reasoning (via planning and logical inference) and deep learning with both the visual stream and the visual search paradigms. 
As a final remark the environment in principle it is not required to be completely static, still vision is egocentric and interaction with people is limited to few actions.  An overview is given in Figure \ref{fig:overview}.

\typeout{--------------------  THE VISUAL STREAM -----------------------}
\section{The visual stream}\label{sec:visualstream}
\begin{figure}
 \centering
\includegraphics[width=8cm,height=6.5cm,]{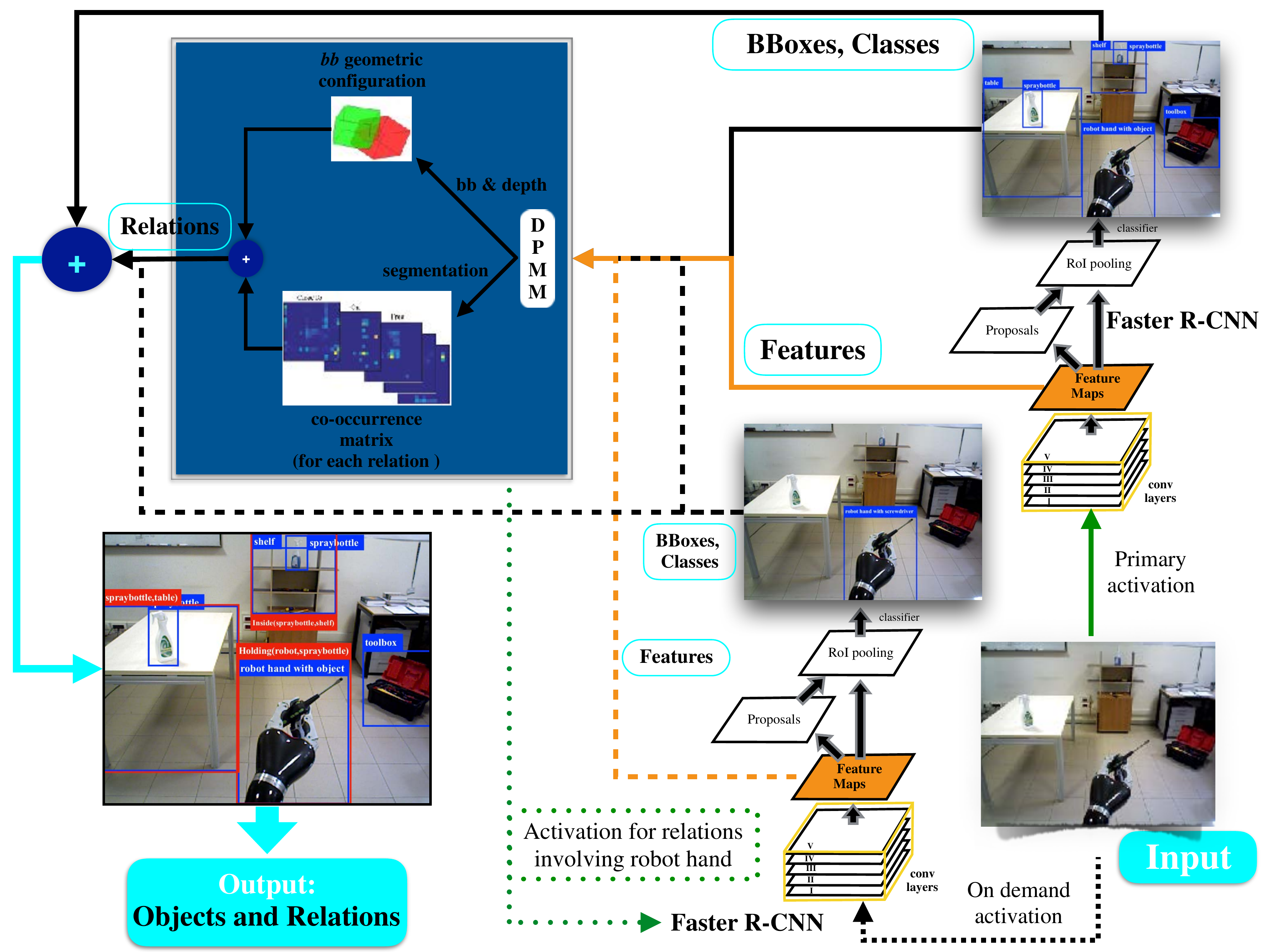}\\
 \caption{The models involved in the visual stream processing for the recognition of the current state $s_t$ concerning preconditions of action $a_{t}$ and postconditions of action $a_{t-1}$.}\label{fig:VS}
\end{figure}

In this section we introduce the visual stream, which is the visual processing of the execution state.  A schema is presented in Figure \ref{fig:VS}. Here a state $s_t$, $t$ in ${\mathbb N}$, specifies the {\em precondition} of action $a_{t}$,  to be executed at state $s_t$, and the {\em postconditions} of action $a_{t-1}$, after its execution. 
For each state $s_t$ of the execution, the VExM opens an input-output channel with the visual stream,   formed by a triple ${\mathcal C}_t = ({\bf u}, F, V)$, where  ${\bf u}$ is a set of objects and relations on them,  expected to hold at execution state $s_t$, $F$ is a buffer, of size $L$, of images from the video recorded by the robot  and $V$ is the  {\em visual stream} processing results, obtained  with a hierarchy of deep models shown in Figure \ref{fig:VS}. 

{\bf The models}. 
There are essentially three models for the visual stream,  which are called in parallel at execution time. 
The first model recognizes elements of the environment, a second model is trained to recognize those items that the robot can handle, and that can be occluded by the hand itself,  a third model   is trained to recognize the spatial relations, some of which are listed in Table \ref{tab:relations}.

{\em Step 1: recognizing objects in the scene}. Let us consider the first two models concerning the objects. The first model is  trained by collecting   images of objects taken from both the robot environment, with an RGBD camera and with images of objects from ImageNet.  The second model is trained with both images of human hands holding objects, and empty, and with the robot hand holding the objects of interest, and empty.  Both these models are based on region proposal networks \cite{ren-2015} and return in $V$ the objects label, their location in the image, and a probability  for the images in $F$, given the $N$ objects in the model. 
Indeed, we are given, for each object in ${\bf u}$:  
\begin{equation}\label{eq:stable}
P(y=ob_j|{\bf x}_{\ell},\varphi_{\ell})= \arg\max_{{\bf x}_{\ell}\in F,\ell\in L}\sum_{\ell=1}^N h({\bf x}_{\ell},\varphi_{\ell})
\end{equation} 
Here $h({\bf x}_{\ell},\varphi_{\ell})$ is the softmax function of the region proposal network \cite{ren-2015} applied to each image in $F$, with $\varphi_{\ell}$ the parameters. 

{\bf Semi-supervised relations estimation}. Despite  recent contributions of several authors on 2d images, there are some difficulties to apply these approaches due to the relevance of depth  to capture spatial relations, from the robot vantage point.  To cope with these difficulties we have defined a  model for relations adapting the active features of the trained DCNN  \cite{ren-2015}, as described above,  to  the RGBD images collected by the robot. For each object $ob_j$ we estimate a statistics of the   active features  with dimension $38\times 50 \times 512$, taken  before the last pooling layer, say $Conv5$, at each pixel inside the recognized object bounding box $bb(ob_j)$ (here we are referring to VGG, though we have considered also ZF, see \cite{zeiler2014,simonyan2014}).   

\begin{figure}
 \centering
  \includegraphics[width=2cm,height=2cm,]{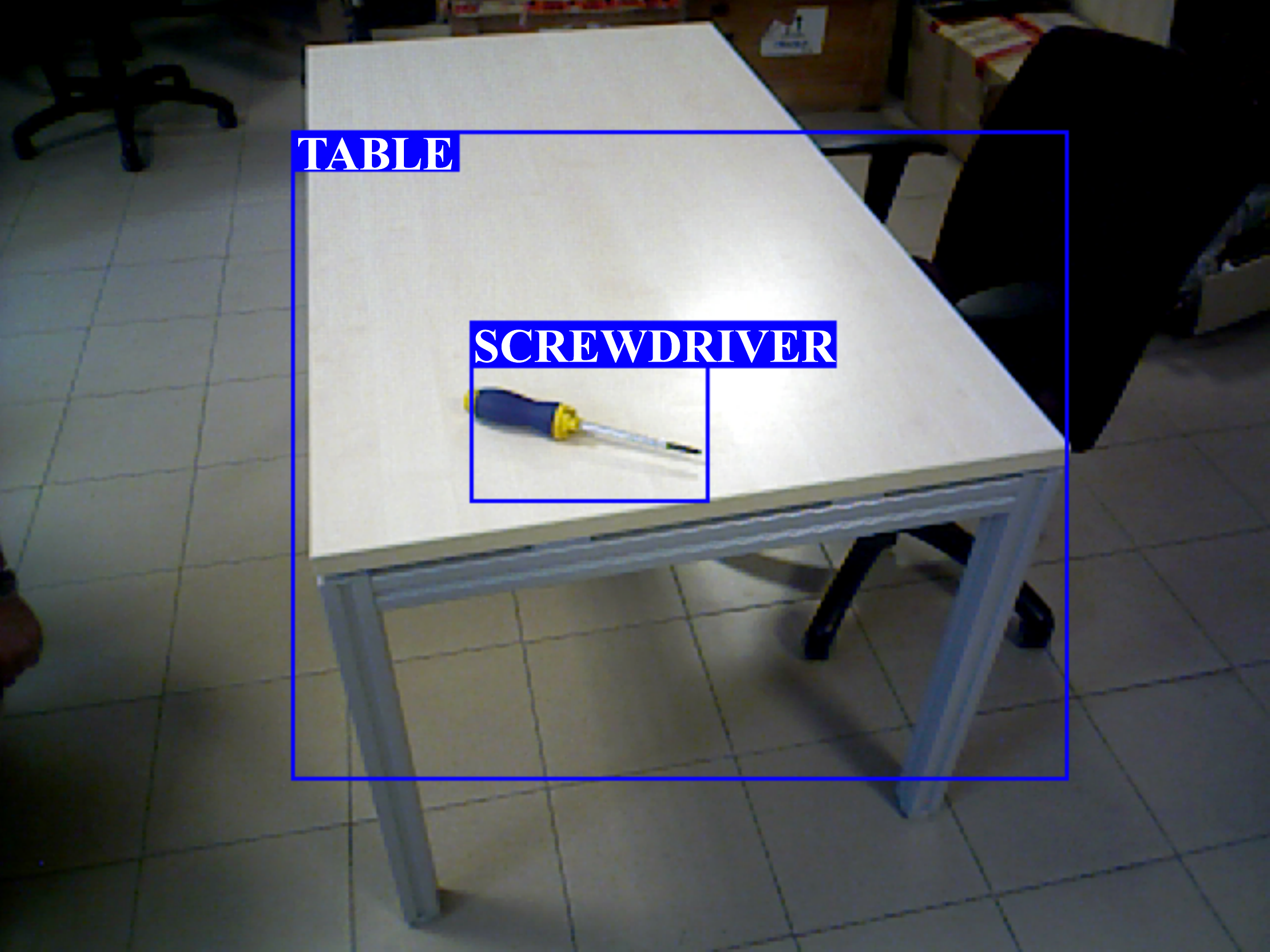}
\includegraphics[width=2cm,height=2cm,]{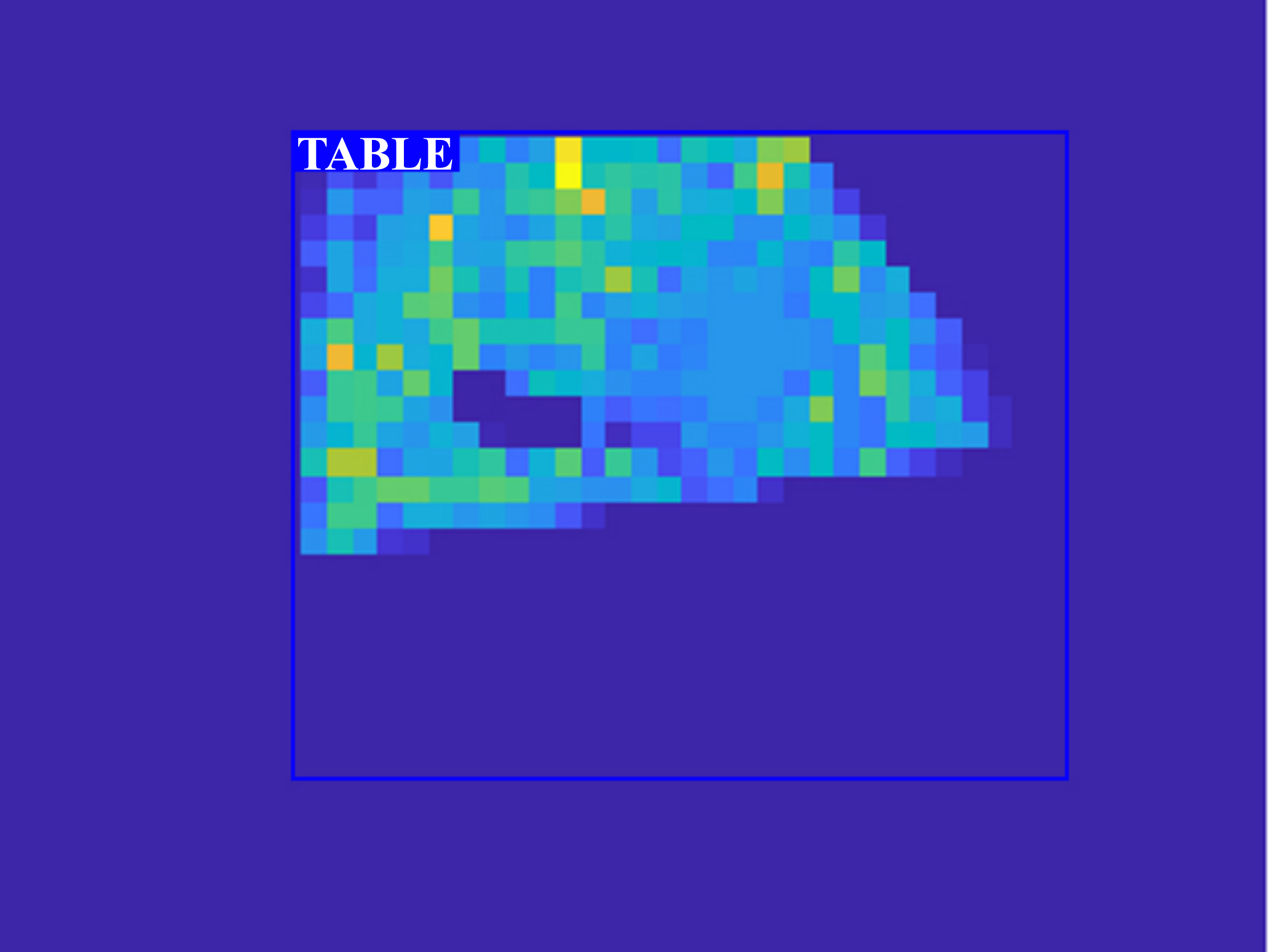}
\includegraphics[width=2cm,height=2cm,]{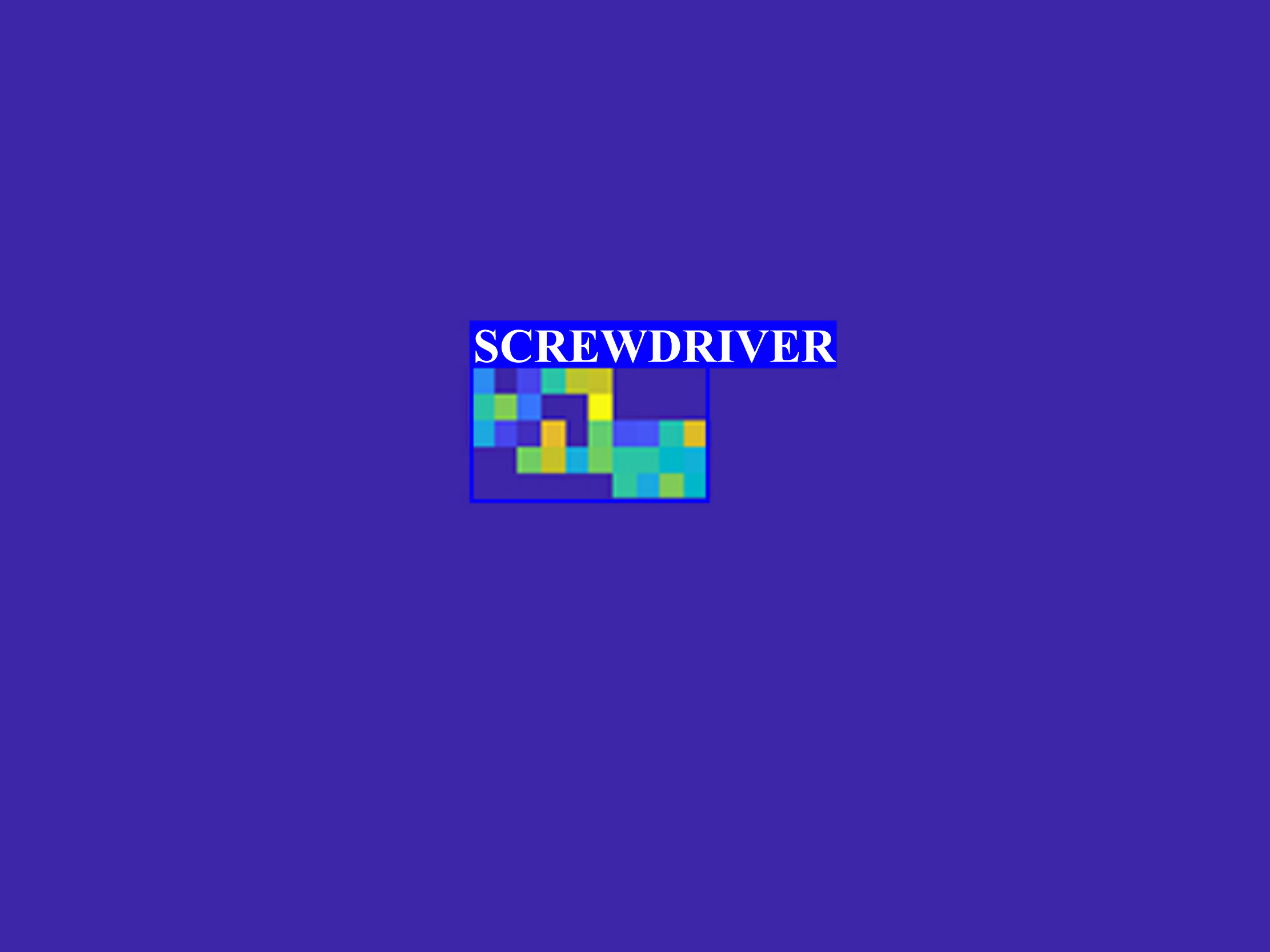}
 \includegraphics[width=2cm,height=2cm,]{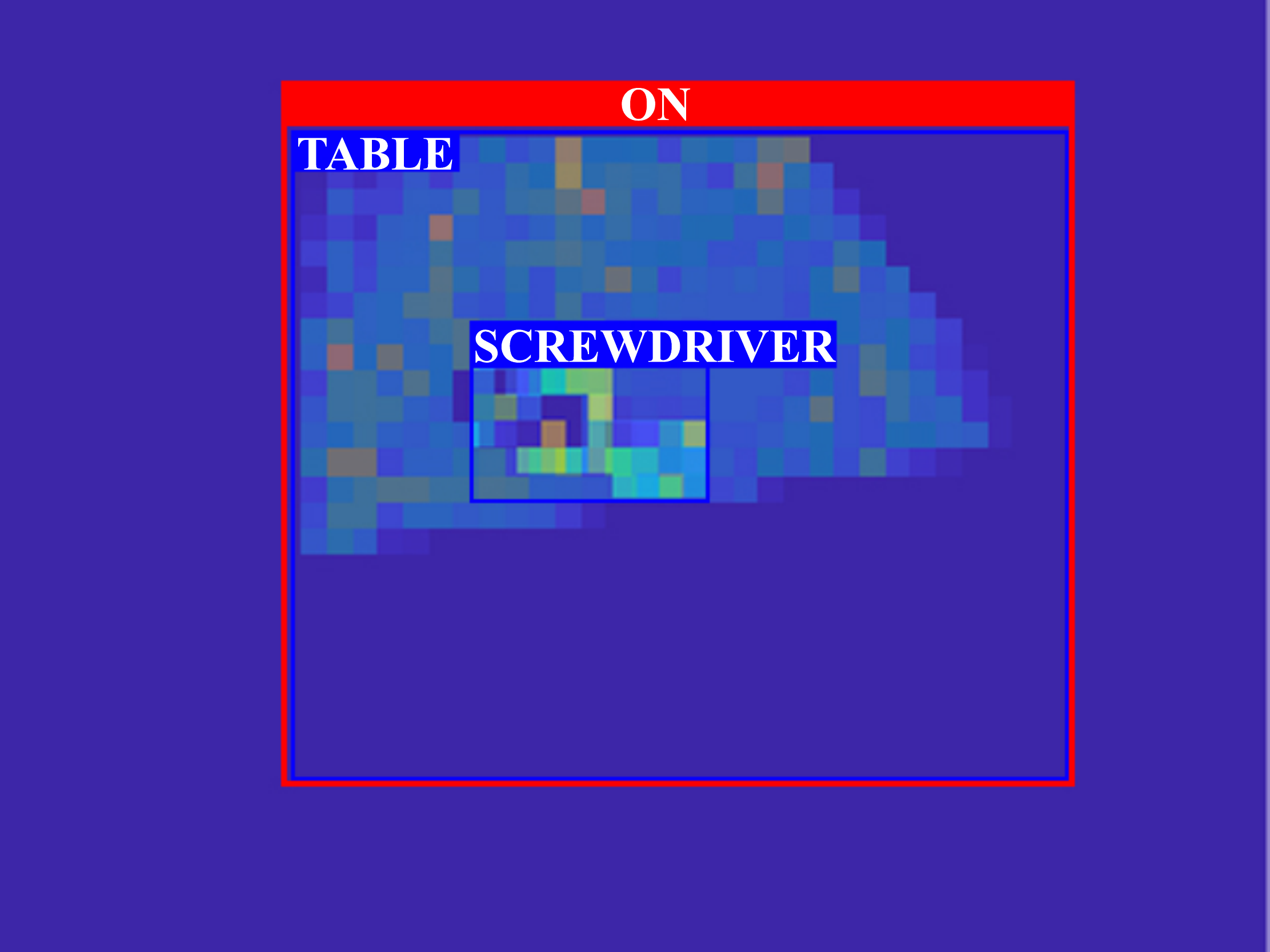}
 \caption{The segmentation map: segmentation of table and screwdriver from the $bb$s in the first image, according to (\ref{eq:seg}), and relation found, according to (\ref{eq:rels}).}\label{fig:segm}

\end{figure}

{\em Step 2: objects segmentation within the bounding boxes}. 
Let the size of $bb(ob_j)$ be $n = u\times v$ and let $z_{ji} = \sum_{k=1}^d x_{jik},i=1,..n$, where $x_{jik}$ is the normalized features value, with $d$ the  feature  dimension  at pixel $i$, indexed w.r.t. the  vectorized region of $bb(ob_j)$.   Let $Z =(z_1, \ldots z_n)^{\top}$ be sorted in descending order,  and let $J$ be the corresponding pixel location of each value in $Z$  within the image resized according to $Conv5$. We consider the vector $\hat{Z}_{ob_j}$ formed by the first 25 elements of $Z$ (hence of $J$), which are the values exceeding the mean square of $Z$, turning out to be inside  the object shape, in the RGBD-image resized according to the size of $Conv5$. Projecting the pixel locations $\hat{J}$ of these values on the resized image we obtain an observation matrix of size $25\times 4$, as we consider the RGB and depth channels. This data collection is repeated for about 100 images for each of the objects. With these features set, for each object we estimate a Dirichlet process mixture (DPM) \cite{ferguson-1973}. Hence we have an infinite mixture   model for each object, which amounts to have a set of parameters $\Theta_h = (\theta_{h,1}, \ldots \theta_{h,k})$ for each object $ob_h$ with mixture components $k=1,..,q$, with $q$ inferred by the process (for more details about the DPM and its computation we refer the interested reader to \cite{teh2011}, see also \cite{sanzari2016,natola2016}). 
Let $ob_{new}$ be an object in ${\bf u}$ - mentioned in the current open channel ${\mathcal C}_t$ - such that for some relation $R_{s}\in {\bf u}$, $ob_{new}$ belongs to the relation domain ${\mathcal D}_{R_s}$.  Then  we seek the model $\Theta_h\in \Theta$ and the component $\theta_{hk}\in \Theta_h$, which maximizes the probability of the active features selected as described above. Namely
\begin{equation}\label{eq:seg}
\begin{array}{l}
M(bb(ob_{new}) | \Theta) = \\
\ \  \ \ \displaystyle{\arg\max_{\theta_{hk}\in \Theta} \{\sum_{h=1}^{N}\sum_{k=1}^q p_{hk} f(\hat{Z} | \theta_{h,k}) \delta_{ob_{new}}({\mathcal D}_{\exists R_s})\}}
\end{array}
\end{equation}
 Note that here $f(\cdot)$ is the normal distribution for which we assumed conjugate priors, $p_{hk}$  are the mixtures weights,  $N$ is  the number of objects, hence of $DPM$  models, and $\delta_{ob_{new}}({\mathcal D}_{\exists R_s})$ is 1 if $ob_{new}\in {\bf u}$ belongs to the domain of some $R_s\in{\bf u}$, and $0$ otherwise. The optimal parameters $\theta_{h,k})$  returns the component model that results in probability map within the  bounding box of object $ob_{new}$, corresponding to  the segmentation for $ob_{new}$, as shown  in Figure \ref{fig:segm} for the table and the screwdriver.  Note that the segmentation is always relative to the region inside the {\em bb}. Note also that  while the DPM model estimation can take several hours, its evaluation  during execution, for the objects in ${\bf u}$, takes a little more than a second, being a second the time needed to run the networks, in order to obtain the active features for all the  objects in ${\bf u}$.

{\em Step 3: computing relations}.
Once the partial segmentations (according to the {\em bb}s) are available, we finally estimate the relations. For each  relation $R_s$  we use basic geometric properties of the spatial configurations  of the two {\em bb}s specifying it, together with  reciprocal depth,  basing on the eight volume-volume relation  model of \cite{egenhofer1995}. Here configurations with the same specification are considered to be topologically equivalent. Let $dmap = g(M(bb(ob_i)|\Theta),M( bb(ob_j) | \Theta'),I_D)$ be the projection of the mask of the probability map  issued by the DPM on the depth map of the  RGBD image in $F$. Let $p_{c}$ be the confidence in the specific $bb$s configuration according to the 9 volume-volume configurations for $dmap$, s.t $sum p_i =1$.  
Without going into further details, we introduce a configuration probability for the $bb$s of the two objects $bb(ob_i)$ and $bb(ob_j)$ as $\nu_{ij}= \beta r^{-1} (A_r Dmap) p_{c}$ with $A_r$ the area of the two $bb$s, $r$ the number of relations and $\beta$ a further normalization constant. 

Configurations of bounding boxes, even including depth, do not convey enough information, since they lack a semantic interpretation of the relations (for example in a specific image {\em Inside} and {\em On} can have the same configuration). Hence  for each relation we define a  co-occurrence matrix $N\times N$ amid all objects. The co-occurrence matrix  reinforces the links between objects that have strong interactions and weakens those links that are already weak. We  estimate the best decomposition  $B = WH$ with $W\in {\mathbb R}_+^{N \times \Lambda}$ and $H \in {\mathbb R}_+^{\Lambda \times N}$, approaching the non-negative factorization (NMF)  basing on the Bayesian NMF proposed in \cite{tan-2013} to infer the best reducing factor $\Lambda$. Let $r$ be the number of relations, considering the tensor resulting from $N \times N \times r$,  link values between $ob_i$ and $ob_j$ along the $r$ coordinate are given by $\lambda_{i,j,r}$. Given these findings,  the probability that the relation $R_{s}$ holds for the observed objects $ob_i, ob_j$ is given by the probability that the geometric configuration of $bb(ob_I)$ and $bb(ob_j)$ and their depth map (according to (\ref{eq:seg})) is correct, and by the likelihood of the relation semantics given  by the co-occurrence matrix refining the relation domain.  Let $\nu$ be the  vector of the configuration values and let $\lambda_{ij}$ be the vector of link values for $ob_i$ and $ob_j$ and denote $z_s(i,j)$ the element wise normalized product of the two vectors. Then the probability of each relation is:
\begin{equation}\label{eq:rels}
 P(R_s | ob_i,ob_j) = z_s(i,j) p(ob_i| \varphi_{\ell})p(ob_j| \varphi_{\ell'}), \ \ \ s=1,\ldots, r
\end{equation}

We see that $z$ acts as a weight and it takes care of both the configuration of the bounding boxes according to the probability map given in  (\ref{eq:seg}) and of the domain of the relations, namely the statistics of their typical appearance in a scene. The chosen relation is the one maximizing the probability. Figure  \ref{fig:segm} show the result for the relation $On(screwdriver,table)$.


\typeout{--------------------  HYBRID PLANNING ENV  -----------------------}
\section{Hybrid planning environment}\label{sec:planenv}

In this section we introduce the hybrid planning environment.  This is made by a deterministic planner and a {\em visual search} policy. The {\em visual search} policy  is required either to focus the robot toward the item of interest, so as to let the visual stream to deal with the preconditions of the action to be executed, or to recover from a failure. In our framework a failure occurs  when the robot misses the objects of interest from the visual field, which is critical since in this case the execution state cannot be verified by the visual stream.    

{\bf The deterministic planner}
The deterministic planner defines a nominal policy of high level actions to the goal.
For the deterministic planner we consider a {\em domain} ${\mathcal D}$  formed by objects, relations and real numbers,  which are represented in the {\em robot language} ${\mathcal L}$.   A term of ${\mathcal L}$ can be an object $ob_i$ of the domain ${\mathcal D}$, a variable or a function $f:{\mathcal D}^n\mapsto {\mathcal D}$, such as actions. Relations (here assumed to be only binary), are defined exclusively for the specified set of objects of the domain, and are interpreted essentially as spatial relations such as {\em On, CloseTo, Inside, Holding}.

A {\em robot task} is defined by a list of plans $\langle {\mathcal P}_1, \ldots, {\mathcal P}_m\rangle$; a {\em plan} ${\mathcal P}_i$ is defined as usual   by an initial state and a set of rules (axioms) specifying the preconditions $H(a_{t})$ and postconditions $K(a_t)$  of  action $a(t)$, where $t$ is discrete time indexed in ${\mathbb N}$. To simplify we assume that preconditions and postconditions are  conjunctions of ground atoms (binary relations), hence a state $s_t {=} H(a_t) \cup K(a_t)$.
Though not necessary here, time  maintains coherence between the deterministic part  of the execution monitor and the Markov decision process underlying the visual search policy computation. 

In our settings robot {\em actions} can be of two types: (1) {\em supply actions}, which are  actions changing the state of the world, in so affecting some object in the world, such as {\em grasp}, {\em hand{-}over}, {\em place}, {\em open}, {\em close}; (2) {\em egocentric actions}, which are actions changing the state of the robot, such as {\em move-Forward},  {\em approach} $ob_j$  along a direction, or {\em look-Up}, {\em look-down}.

Egocentric actions, despite belong to the deterministic planner language, are mostly used within  learned {\em visual search}  policies, which are inferred according to the current state given a deep RL model \cite{mnih2016}. On the other hand supply actions can have non-deterministic effects, which then lead to a failure, handled by a visual search recover policy.

According to these definitions a {\em robot plan} specifies at most one supply-action  and this action is a {\em durable action},  defined by an action {\em start} and by an action {\em end}. Any other action  in the plan is an egocentric action.  In particular, given the initial state, each plan introduces at most a new  primal object  the robot can deal with and the relations of this object with the domain. An example is as follows:  robot is holding an object, as a result of a previous plan, and it has to put it away on a toolbox. This ensures that recovery from a failure is circumscribed for each plan to a single supply action.

The {\em task goal} is unpacked into an ordered list of goals ${\mathcal G}{=}(G_1,\ldots, G_m)$ such that each $G_i$ is the goal of plan ${\mathcal P}_i$ in the list of plans forming a task. 
Given a goal $G_i$, if a  plan ${\mathcal P}_i$ leading from the initial state to the goal exists, then a sequence of actions leading to the goal is inferred by the search engine \cite{helmert-2006}.  Here plans are defined  in PDDL \cite{edelkamp-2004}. 

We extend \cite{helmert-2006} parser to inferred plans, so as to obtain the initial state, the sequence of actions $(a_0, \ldots, a_n)$ leading to the goal, and the set  of states $S_{{\mathcal P}_i}$ for each plan ${\mathcal P}_i$. This list  forms the set of deterministic states and transitions. 

In summary, the {\em deterministic planner plays the role of an owner's manual} providing the robot with  instructions about  actions,  properties,  conditions that in principle   should hold to accomplish the task.

{\bf The visual search Policy}. Given the visual stream and the definition of a state we make here the assumption that the environment is fully observable. Hence we introduce a simple hybrid MDP, with discrete time, formed by the quintuple $(S,A,T,RW,\gamma)_{\mathcal P}$, for each plan ${\mathcal P}$.  Here $S,A$ (set of states and actions for ${\mathcal P}$) are  generated by the plan parser, $T:S\times A\times S\mapsto {\mathbb R}$ is the transition probability distribution, specifying the probability of moving between states, namely each $T_{ij} {=}\tau(s_{t+1} | s_t,a_t)$. $T$ has an added {\em failure state}. Initially $T$ (which is relative to a plan ${\mathcal P_i}$ hence should be written $T_{{\mathcal P}_i}$) is generated by the parser as a right stochastic $0{-}1$ matrix, and further updated by the VExM (see next section); $RW: S\mapsto {\mathbb R}$ is the expected reward and $\gamma\in [0,1]$ is the discount factor. In particular $T= \kappa (T_{pre}+T_{post})$, since states transitions are checked and updated by the VExM before and after action execution.

We recall here some notions from MDP and RL \cite{bertsekas1995,sutton1998}.  The value $v$ of a state $s$ is the expected return starting from state $s_t$, namely $v(s){=}E[\sum_{k{=}0} \gamma^k RW_{t+k} | s_t{=}s]$, in matrix form (for each state) ${\bf v} {=} RW + \gamma T {\bf v}$. A policy $\pi: S\mapsto A$ specifies  the probability of an action $a\in A$ given state $s\in S$, it is time-independent  and has the Markov property. The action value function $Q_{\pi} (s,a)$ simply extends $v(s)$ to $v(s,a)$. The optimal  action-value function $Q^{*}(s,a){=} \max_{\pi} Q_{\pi}(s,a)$ is the maximum action-value function over all policies  and an optimal policy is one that achieves the optimal action-value function $\pi^{\star}(a,s) {=} \arg\max_{a} Q^{\star}(a,s)$.  

We are interested in an optimal visual search policy, which is a sequence of egocentric actions leading the robot to observe the scene so that the preconditions of the action to be executed or the postconditions of the executed action can be observed and verified. So the search policy drives the visual stream, which cannot be done by the blind deterministic planner.
As noted above visual search can be composed of the following actions: {\em look-Left, look-Right, look-Up, look-Down, move-Forward, move-Backward, turn-Left, turn-Right}, with all these actions specified by precise quantities coherent with the underlying control, not mentioned here.

We approach the problem with a policy-based method by computing the parametrized policy $\pi(a| s,\psi)$ \cite{williams1992}, where $\psi$ is the policy weight vector, hence $\pi(a|s,\psi){=} P(A_t {=}a|s_t{=}s,\psi_t{=}\psi)$, and the parameter update is:
\begin{equation}\label{eq:pu}
\begin{array}{l}
\psi_{t+1} {=}  \psi_t+ \\
\ \ \ \alpha\left(RW_{t+1}+\gamma \hat{v}(s_{t+1},{\bf w})-\hat{v}(s_t,{\bf w})\right)  \nabla_{\psi}\log \pi(a_t| s_t,\psi) 
\end{array}
\end{equation}
where $\hat{v}$ is a differentiable state value parametrization, ${\bf w}\in {\mathbb R}^m$ are state-value weights,  $\psi$ the policy weights, and  $\alpha>0$ step size parameter \cite{sutton1998}. This is an actor-critic model in which the critic learns the action-value function while the actor learns the policy $\pi$. The goal is to learn a visual search policy.
With A3C \cite{mnih2016} have shown  that a DNN can be used to approximate both the policy $\pi(a_t| s_t,\psi)$ (softmax over actions) and the value function using two cost functions, the actor $L_{\pi}(\psi)$, which is derived from (\ref{eq:pu}) with an extra entropy term to boost exploration in training, and the critic  $L_{v}(\psi){=} (\sum_{i{=}0}\gamma^i RW_{t+i} -v(s_t | \psi)^2$.  For optimization they used the standard non-centered RMSProp update: $g {=} \alpha g +(1-\alpha)\Delta \psi^2$, and $\psi \leftarrow \psi -\eta \Delta\psi(g+\epsilon)^{-1/2}$.  Training has been done with A3C-LSTM  \cite{mnih2016}, though with input the mental maps, described in the following. Our implementation follows \url{https://github.com/miyosuda/async_deep_reinforce}. A reward of 1 is issued whenever the primal object of the current plan appears in the mental map, then the robot restart the experiment.

\begin{figure}
 \centering
  	\includegraphics[width=4cm,height=2.5cm,]{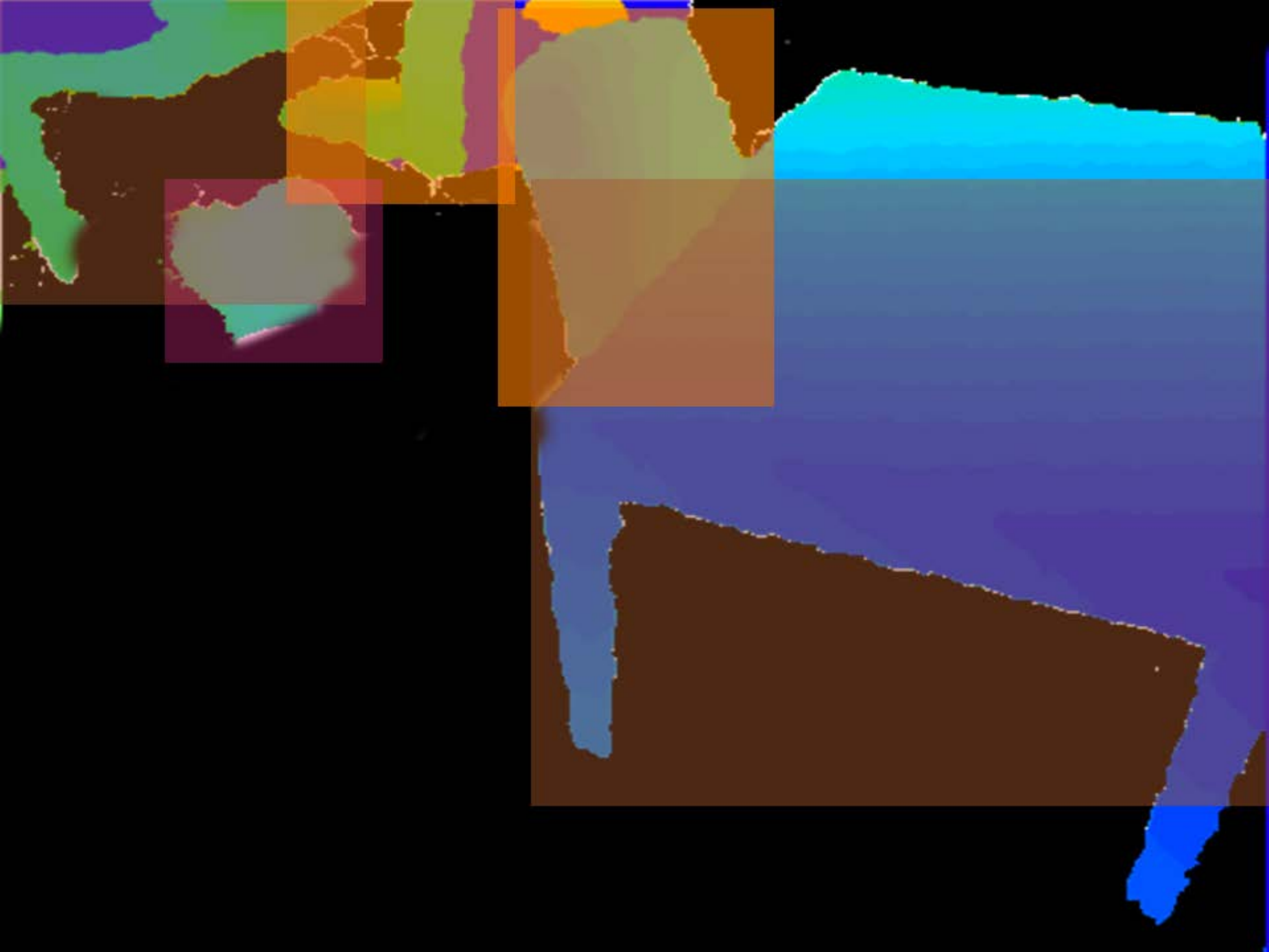}
	\includegraphics[width=4cm,height=2.5cm,]{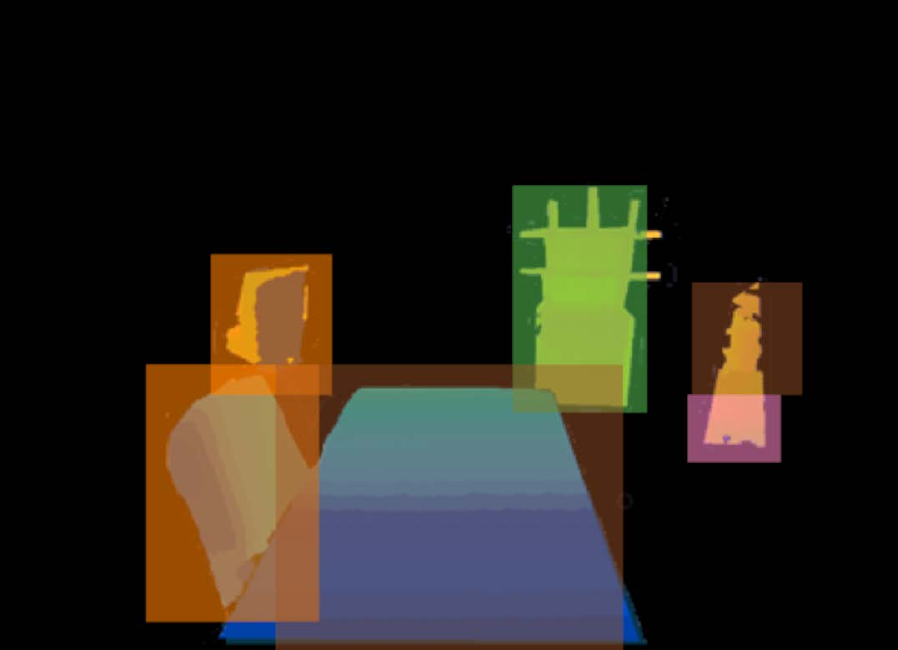}\\
	\includegraphics[width=4cm,height=2.5cm,]{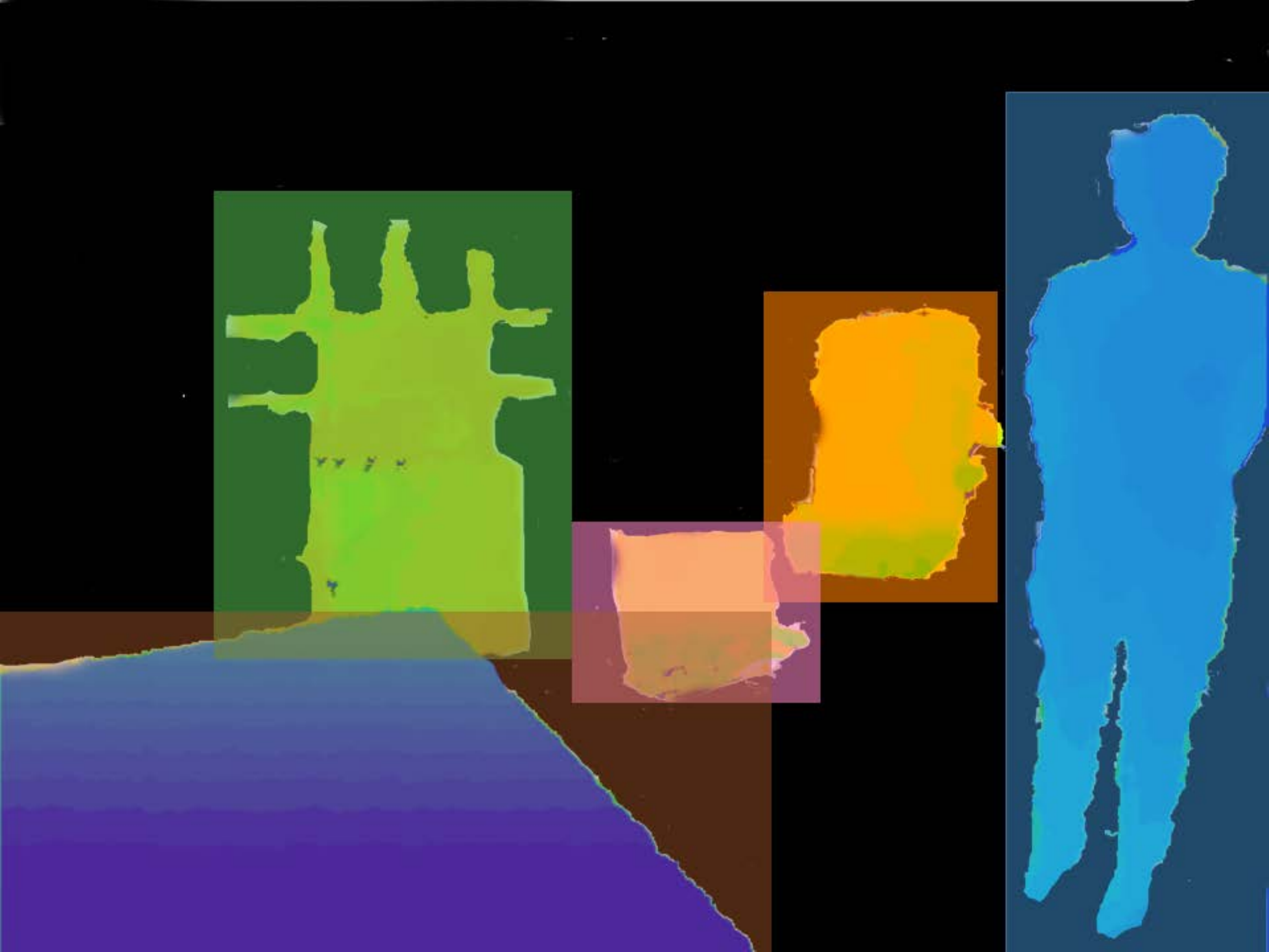}
	 \includegraphics[width=4cm,height=2.5cm,]{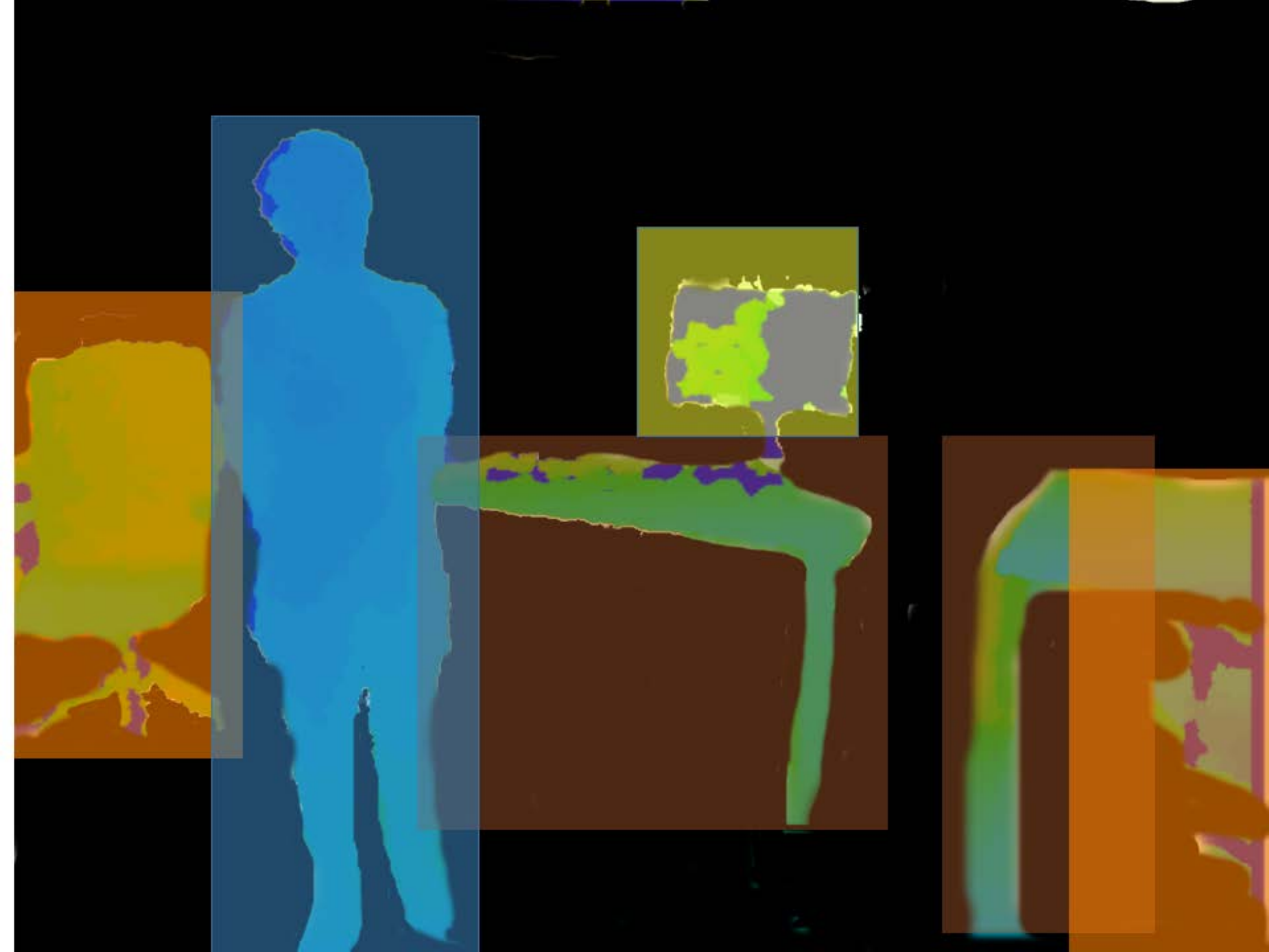}
 \caption{Mental Map generation for searching a person. The transparency shows the depth map of the labeled objects}\label{fig:MM}
\end{figure}

{\bf Mental Maps} There is an increasing interest in creating environment for deep RL, taking raw pixels as input, for example  \cite{mottaghi2016,ZhuGKFFGMF17,zhu2017}.  Here we use as input   to the DNN the automatically generated mental maps, collected during experiments. The mental map takes as input an image from $F$ labeled by $V$ with stable bounding boxes $bb$s according to (\ref{eq:stable}), the depth of the segmented objects within the bbs and a color code for each of the $N$ objects in the language and a link color, represented by a colored line, between the objects in a relation. It then forms a colored map which represents everything the robot knows about the world. In  Figure \ref{fig:MM} we show a short sequence of close mental maps generated for searching a person. 

\typeout{--------------------  FULL EX - MONITORING -----------------------}

\section{Full execution algorithm}\label{sec:monitor}
In this section we introduce the VExM algorithm.  In the current implementation, given a task, the VExM is able to choose the list of goals on the basis of a naive Bayes classifier taking as input a bag of words, which we have not described here.
Given the goals the choice of a deterministic plan for each goal is immediate. 
For each plan the VExM compute the parser, which in turns infers the ordered list of egocentric and supply actions $A$ and states $S$ to obtain the goal. We recall that a state $S_t$ specifies the set of preconditions and postconditions of action $a_t$, namely $H(a_t) \cup K(a_t)$.
The algorithm starts with the ordered list of actions and states as computed by the parser for a plan ${\mathcal P}_i$ in the list of plans, and ends with the list of executed actions for ${\mathcal P}_i$, where only one action is a {\em supply action}. Note also that several events are here  not taken into account, such as camera malfunctioning or other events concerning control and navigation that cannot be currently handled by the high-level VExM. 

At each state $s_t$ the VExM computes via the visual stream the probability of each relation in ${\bf u}$, as described in Section \ref{sec:visualstream} and a loss. Then it updates the transition $T$, starting in $s_t$ with the truth values inferred by the deterministic planner. Because the monitor evaluates the loss for both preconditions and postconditions separately we indicate two loss functions, namely ${\mathcal L}(s_t,X(a_t))$, with $X\in\{H,K\}$. Given $a_t$ at state $s_t$ and for  $R_s\in {\bf u}_t$, $r$ the number of relations:
\begin{equation}\label{eq:loss}
\begin{array}{ll}
1.\ {\mathcal L}(s_t,X(a_t)) = \\
 \ \ \ \ \displaystyle{ \frac{1}{r} \sum_{R_s\in {\bf u}_{H(a_t)}} \| 1 - P(R_s(ob_i,ob_j))\| \delta_{ob_i}(R_s)\delta_{ob_j}(R_s)}
\end{array}
\end{equation}
Here $P(R_s(ob_i,ob_j))$ is given in (\ref{eq:rels}). 
According to (\ref{eq:loss})  the transition matrix between states $s_t$ and $s_{t+1}$, is updated  $T'_{pre} = \left({\mathcal L}(s_t,H(a_t))T_{pre}\right) /\left(2\sum_{s\in S} {\mathcal L}(s_t,H(a_t))T_{pre}\right)$ and similarly for $T_{post}$, hence $T = \kappa(T_{pre}+T_{post})$. This include transition $\tau(s_{failure}| s_t,a_t)$ between state $s_t$ and the added failure state  $s_{failure}$,(see Section \ref{sec:planenv}), so that the sum is always one. If the transition value to the failure state is greater than the transition value to the next state the visual search policy  is called, independently of the planning requests. The algorithms are shown below. 

\begin{algorithm}%
\caption{Execution Monitor.} \label{algo:algo1}%
\KwIn{Ordered list $LI$ of actions and states for plan ${\mathcal P}_i$}%
\KwOut{List $LE$ of experienced actions}%
\While{$LI\neq []$} {%
    Choose $a_t\in LI$, $t= 0, \ldots, n$\;
    Open $({\bf u}, F, V)_t$ with {\em visual stream}\;
    Compute $V_{H(a_t)}$ and update transition $T_{pre}$ according to (\ref{eq:loss}) for $H$\;%
    Compute Mmaps\;
    \lIf {$a_t$ is visual search and args = $ob_j$}
{
    $[\widehat{{\bf u}}_t, \widehat{F}_t, \widehat{V}_t,\pi(\dot),\reallywidehat{Mmaps}_t]$ = 
         $Vsearch_{\pi}(a_t,ob_j,({\bf u}, F, V)_t,Mmaps)$\;
   $\vec{a_t} \leftarrow \pi(s_t), {\bf u}\leftarrow \widehat{\bf u}_t, F\leftarrow \widehat{F}_t$, 
     $V_t\leftarrow \widehat{V}_t, Mmaps = Mmaps\cup\{\reallywidehat{Mmaps}_t\}$\;
     Execute $\vec{a_t}$ involving navigation and control, if required\;
     Compute $V_{K(a_t)}$ and update transition $T_{post}$ according to (\ref{eq:loss}) for $K$\;%
    $LE = LE\cup\{\vec{a_t}\}$
}
    \uElseIf{$\tau(s_{failure}| s_t,a_t) <\tau(s_{t+1}|s_t,a_t)$ or $a_t$ is not supply-action}
        { 
           Execute action $a_t$ involving navigation and control, if required\;
           Compute $V_{K(a_t)}$ and update transition $T_{post}$ according to (\ref{eq:loss}) for $K$\;
          $LE = LE\cup\{\vec{a_t}\}$
        }
           \ElseIf{$\tau(s_{failure}| s_t, a_t) >\tau(s_{t+1}|s_t,a_t)$ and $a_t$ is  supply-action} 
{               $[\widehat{{\bf u}}_t, \widehat{F}_t, \widehat{V}_t,\pi(\dot),\reallywidehat{Mmaps}_t]$ = 
         $Vsearch_{\pi}(a_t,ob_j,({\bf u}, F, V)_t,Mmaps)$\;
   $\vec{a_t} \leftarrow \pi(s_t), {\bf u}\leftarrow \widehat{\bf u}_t, F\leftarrow \widehat{F}_t$, 
     $V_t\leftarrow \widehat{V}_t, Mmaps = Mmaps\cup\{\reallywidehat{Mmaps}_t\}$\;
             
               Execute $\vec{a_t}$ involving navigation and control, if required\;
               Compute $V_{K(a_t)}$ and update transition $T_{post}$ according to (\ref{eq:loss}) for $K$\;%
    $LE = LE\cup\{\vec{a_t}\}$
  }
       $T =\kappa(T_{pre}+T_{post})$\;
      $t\leftarrow t+1$\;
}
\eIf{$Empty$ list of actions}{\Return $LE$}{Failure}
\end{algorithm}%

\begin{algorithm}%
\caption{Visual Search $Vsearch$} \label{algo:algo2}%
\KwIn{$(a_t,ob_j,({\bf u}, F, V)_t,Mmaps)$ }%
\KwOut{ $[\hat{{\bf u}}_t, \hat{F}_t, \hat{V}_t,\pi_t,\reallywidehat{Mmaps}_t]$}%
  Compute $\pi(s_t)$, for visual search $ob_j$, with $ob_j$ argument of $a_t$, input $Mmaps$, with A3C model\;
  Compute $\reallywidehat{Mmaps}$\;
  \Return $\widehat{{\bf u}}_t, \widehat{F}_t, \widehat{V}_t$, $\pi$
 \end{algorithm}%

\typeout{-------------------- EXPERIMENTS -----------------------}

\section{Experiments and results}\label{sec:experiments}
\noindent
{\bf Platform}
Experiments for the VExM presented here have been done under different conditions in order to test different aspects of the model. To begin with, all experiments have been performed with a custom-made robot.  A Pioneer 3 DX differential-drive robot is used as a compact mobile base. To interact with the environment we mounted a Kinova Jaco 2 arm with 6 degrees of freedom and a reach of 900mm and finally, for visual perception, we used an Asus Xtion PRO live RGB-D camera mounted on a Direct Perception PTU-46-17.5 pan-tilt unit. 

\noindent
{\bf Tasks}
We have considered two classes of tasks: (1) bring an object (spraybottle, screwdriver,hammer,cup) on the table or inside the shelf to a subject;  (2) put-away object (screwdriver, hammer, spanner) in the toolbox. Each experiment in a class has been run 35 times manually driving the robot to collect images of the scene and 20 more times with the planning environment described in Section \ref{sec:planenv}, despite in a number of circumstances the grasping action has been manually helped, especially with small objects. 
From robot experiments we collected 120000 images, while from ImageNet we collected 25000 images. Table \ref{tab:relations} shows the main relations, objects and actions considered in the tasks.

\begin{table}[ht!]
\caption{Subset of Objects, relations, supply-actions and egocentric actions of the robot language ${\mathcal L}$}\label{tab:relations}
\centering
\resizebox{0.98\columnwidth}{!}{
\begin{tabular}{l|l|l|l}
\textbf{\small Relations} & \textbf{\small Objects} & \textbf{\small Supply actions} & \textbf{\small Egocentric actions} \\ \hline

\small CloseTo & \small Bottle & \small Close & \small Look-down \\

\small Found & \small Chair & \small Grasp & \small Look-left \\

\small Free & \small Cup & \small Open & \small Look-right \\

\small Holding & \small Floor & \small Hand-over & \small Look-up\\

\small Inside & \small Hammer & \small Place &  \small Move-forward \\

\small On & \small Person & \small Lift &  \small Move-backward\\

\small InFront& \small Spray Bottle & \small Push  & \small Turn-left\\

\small Left & \small Screwdriver &\small Spin & \small Turn-right\\

\small Right & \small Shelf & Dispose  & \small Localize \\

\small Under & \small Toolbox  & & \small Rise-arm \\

\small Behind & \small TV-Monitor & &   \small Lower-arm \\

\small Clear & \small Table &  & \small Close-Hand\\
\small Empty & \small Door & & \small Open-Hand\\
\end{tabular}}
\end{table}

\noindent
{\bf Training}
We train the DCNN models using images taken from the ImageNet dataset, as well as images collected by the ASUS Xtion PRO RGB-D camera. We split the set of images in training and validation sets with a proportion of 80\%-20\%. We performed 70000 training iterations for each model on a PC equipped with 8 GPUs. Table II shows the object detection accuracy achieved by the two DCNN models dealing with the free and holding settings, respectively.

\setcounter{table}{2}

\begin{table}[ht!]
\centering
\caption{Relations detection accuracy}\label{tab:accuracy}
\resizebox{0.5\columnwidth}{!}{
\begin{tabular}{l|c}
\textbf{\small Relations} & \textbf{\small Accuracy}\\ \hline
\small CloseTo & 64\% \\
\small Found & 73\% \\
\small Free & 62\%   \\
\small Holding & 79\% \\
\small Inside & 75\%   \\
\small On & 83\% \\]
\small InFront & 78\%  \\
\small Left & 68\%  \\
\small Right  & 72\% \\
\small Under  & 71\% \\
\small Behind & 89\% \\
\small Clear & 67\% \\
\small Empty & 76\%  \\ \hline
\textbf{\small Average} & 73.6\%
\end{tabular}
}
\end{table}

\begin{figure}[ht!]
 \centering
  \includegraphics[width=0.95\columnwidth]{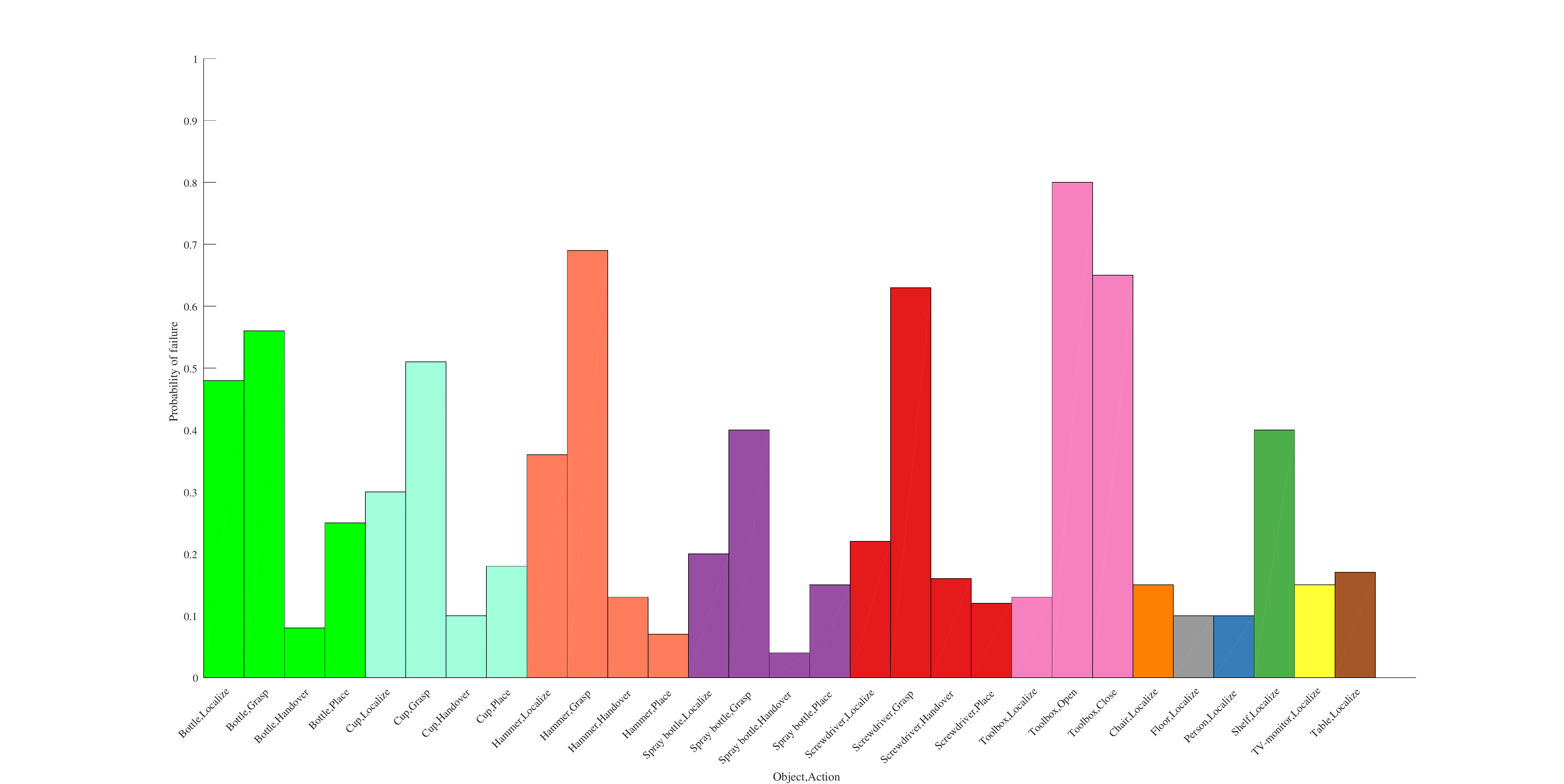}\\
 \caption{Histogram of action/objects failures}\label{fig:histo1}
\end{figure}

\noindent
{\bf Failures}
We examine the number of failures encountered during the executions of the tasks described above. A failure is recorded as soon as the state perceived by the visual stream via the DCNN and DPM models, does not match the post-conditions of the action executed. The histogram in Fig. \ref{fig:histo1} shows the probability that a failure is encounter while executing a particular action  in relation to the object being involved. We note that, as expected, more complex actions like grasping and localizing show a higher probability of failure. Surprisingly, handing over an item shows a low failure probability, this is mainly attributed to the high adaptability of the subject involved in the action.

\begin{figure}[ht!]
 \centering
  \includegraphics[width=0.8\columnwidth ,height=4.5cm]{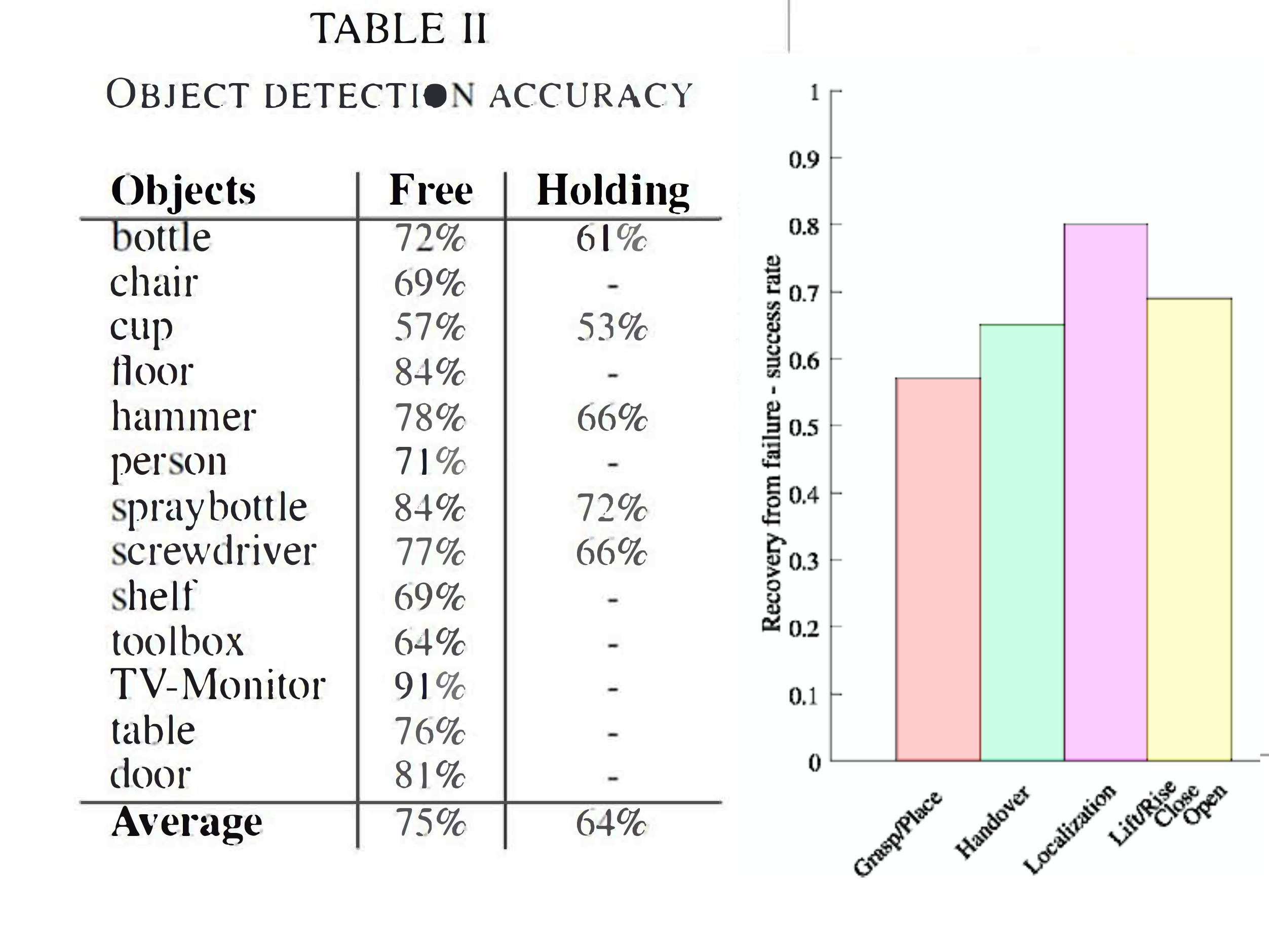}\\
 \caption{On the right the histogram of the recovery success ratios via visual search policy}\label{fig:histo2}
\end{figure}

\noindent
{\bf Recovery}
As explained in Section \ref{sec:planenv}, a visual search policy is employed as soon as a failure is detected in order to find the primary objects involved in the action that failed. Fig. \ref{fig:histo2} shows the success rate of the recovery via the use of a visual search policy for four different types of actions. Localization and manipulation actions show a higher recovery success rate, while recovering from a failed grasping seems to be the most problematic.

\section*{ACKNOWLEDGMENT}
This research is supported by EU H2020 Project Secondhands 643950.

\bibliographystyle{IEEEtran}
\bibliography{bibliography}

\end{document}